# Weighted Automata in Text and Speech Processing

**Mehryar Mohri, Fernando Pereira, Michael Riley**
**AT&T Research**
**600 Mountain Avenue**
**Murray Hill, 07974 NJ**
{mohri,pereira,riley}@research.att.com

**Abstract.** Finite-state automata are a very effective tool in natural language processing. However, in a variety of applications and especially in speech precessing, it is necessary to consider more general machines in which arcs are assigned weights or costs. We briefly describe some of the main theoretical and algorithmic aspects of these machines. In particular, we describe an efficient composition algorithm for weighted transducers, and give examples illustrating the value of determinization and minimization algorithms for weighted automata.

## 1 Introduction

Finite-state acceptors and transducers have been successfully used in many natural language-processing applications, for instance the compilation of morphological and phonological rules [5, 10] and the compact representation of very large dictionaries [8, 12]. An important reason for those successes is that complex acceptors and transducers can be conveniently built from simpler ones by using a standard set of algebraic operations — the standard rational operations together with transducer composition — that can be efficiently implemented [4, 2]. However, applications such as speech processing require the use of more general devices: weighted acceptors and weighted transducers, that is, automata in which transitions are assigned a weight in addition to the usual transition labels. We briefly sketch here the main theoretical and algorithmic aspects of weighted acceptors and transducers and their application to speech processing. Our novel contributions include a general way of representing recognition tasks with weighted transducers, the use of transducers to represent context-dependencies in recognition, efficient algorithms for on-the-fly composition of weighted transducers, and efficient algorithms for determinizing and minimizing weighted automata, including an on-the-fly determinization algorithm to remove redundancies in the interface between a recognizer and subsequent language processing.

## 2 Speech processing

In our work we use weighted automata as a simple and efficient representation for all the inputs, outputs and domain information in speech recognition above the signal processing level. In particular, we use transducer composition to represent the combination of the various levels of acoustic, phonetic and linguistic information used in a recognizer [11]. For example, we may decompose a recognition task into a weighted acceptor $O$ describing the acoustic observation sequence for the utterance to be recognized, a transducer $A$ mapping acoustic observation sequences to context-dependent phone sequences, a transducer $C$ that converts between sequences of context-dependent and context-independent phonetic units, a transducer $D$ from context-independent unit sequences to word sequences and a weighted acceptor $M$ that specifies the *language model*, that is, the likelihoods of different lexical transcriptions (Figure 2).

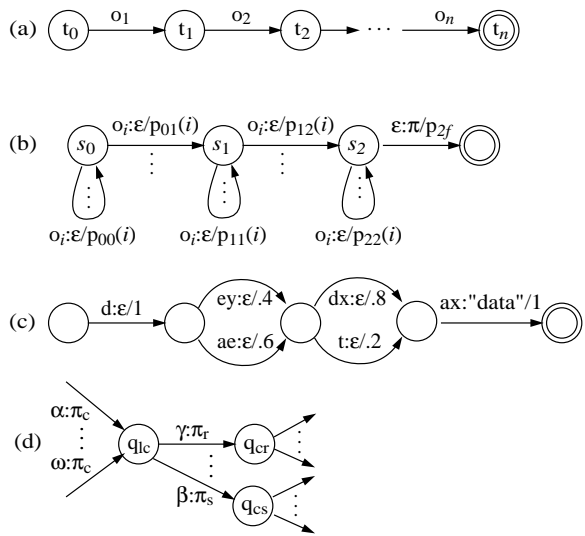

**Figure 1.** Models as Weighted Transducers.

The trivial acoustic observation acceptor $O$ for the vector-quantized representation of a given utterance is depicted in Figure 1a. Each state represents a point in time $t_i$, and the transition from $t_{i-1}$ to $t_i$ is labeled with the name $o_i$ of the quantization cell that contains the acoustic parameter vector for the sample at time $t_{i-1}$. For continuous-density acoustic representations, there would be a transition from $t_{i-1}$ to $t_i$ labeled with a distribution name and the likelihood[1] of that distribution generating the acoustic-parameter vector, for each acoustic-parameter distribution in the acoustic model.

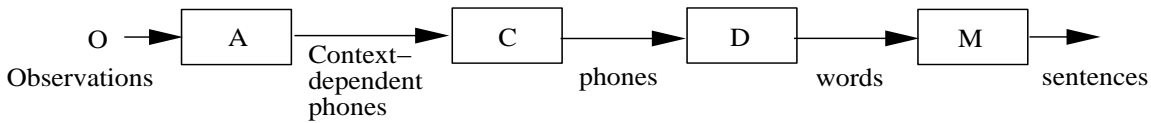

**Figure 2.** Recognition Cascade

The acoustic-observation sequence to phone sequence transducer $A$ is built from *context-dependent* (CD) phone models. A CD-phone model is given as a transducer from a sequence of acoustic observation labels to a specific context-dependent unit, and assigns to each acoustic sequence the likelihood that the specified unit produced it. Thus, different paths through a CD-phone model correspond to different acoustic realizations of a CD-phone. Figure 1b depicts a common topology for such a CD-phone model. The full acoustic-to-phone transducer $A$ is then defined by an appropriate algebraic combination (Kleene closure of sum) of CD-phone models.

The form of $C$ for triphonic CD-models is depicted in Figure 1d. For each context-dependent phone model $\gamma$, which corresponds to the (context-independent) phone $\pi_c$ in the context of $\pi_l$ and $\pi_r$, there is a state $q_{lc}$ in $C$ for the biphone $\pi_l\pi_c$, a state $q_{cr}$ for $\pi_c\pi_r$ and a transition from $q_{lc}$ to $q_{cr}$ with input label $\gamma$ and output label $\pi_r$. We have used transducers of that and related forms to map context-independent phonetic representations to context-dependent representations for a variety of medium to large vocabulary speech recognition tasks. We are thus able to provide full-context dependency, with its well-known accuracy advantages [7], without having to build special-purpose context-dependency machinery in the recognizer.

The transducer $D$ from phone sequences to word sequences is defined similarly to $A$. We first build *word models* as transducers from sequences of phone labels to a specific word, which assign to each phone sequence a likelihood that the specified word produced it. Thus, different paths through a word model correspond to different phonetic realizations of the word. Figure 1c shows a typical word model. $D$ is then defined as an appropriate algebraic combination of word models.

Finally, the language model $M$, which may be for example an $n$-gram model of word sequence statistics, is easily represented as a weighted acceptor.

The overall recognition task can be then expressed as the search for the highest-likelihood string in the composition $O \circ A \circ C \circ D \circ M$ of the various transducers just described, which is an acceptor assigning each word sequence the likelihood that it could have generated the given acoustic observations. For efficiency, we use the standard Viterbi approximation and search for the highest-probability path rather than the highest-probability string.

Finite-state models have been used widely in speech recognition for quite a while, in the form of hidden-Markov models and related probabilistic automata for acoustic modeling [1] and of various probabilistic language models. However, our approach of expressing the recognition task with transducer composition and of representing context dependencies with transducers allows a new flexibility and greater uniformity and modularity in building and optimizing recognizers.

## 3 Theoretical definitions

In a recognition cascade such as the one we just discussed (Figure 2), each step implements a mapping from input-output pairs $(r, s)$ to probabilities $P(s|r)$. More formally, each transducer in the cascade implements a *weighted transduction*. A weighted transduction $T$ is a mapping $T : \Sigma^* \times \Gamma^* \to K$ where $\Sigma^*$ and $\Gamma^*$ are the sets of strings over the alphabets $\Sigma$ and $\Gamma$, and $K$ is an appropriate weight structure, for instance the real numbers between 0 and 1 in the case of probabilities[2]

The right-most step of (Figure 2), the language model acceptor $M$, represents not a transduction but a *weighted language*. However, we can identify any weighted language $L$ with the restriction of the identity transduction that assigns to $(w, w)$ the same weight as $L$ assigns to $w$, and in the rest of the paper we will use this to identify languages with transductions and acceptors with transducers as appropriate.

Given two transductions $S : \Sigma^* \times \Gamma^* \to K$ and $T : \Gamma^* \times \Delta^* \to K$, we can define their *composition* $S \circ T$ by

$$(S \circ T)(r, t) = \sum_{s \in \Gamma^*} S(r, s)T(s, t) \quad (1)$$

For example, if $S$ represents $P(s_l|s_i)$ and $T$ $P(s_j|s_l)$ in (2), $S \circ T$ represents $P(s_j|s_i)$.

It is easy to see that composition $\circ$ is associative, that is, in any transduction cascade $R_1 \circ \cdots \circ R_m$, the order of association of the $\circ$ operators does not matter.

*Rational* weighted transductions and languages are those that can be defined by application of appropriate generalizations of the standard Kleene operations (union-sum, concatenation and closure) and are also exactly those implemented by weighted finite-state automata (transducers and acceptors) [4, 3, 6]. We are thus justified in our abuse of language in using the same terms and symbols for rational transduction and finite-state transducer operations in what follows.

---

[1] For computational reasons, sums and products of probabilities are often replaced by minimizations and sums of negative log probabilities. Formally, this corresponds to changing the *semiring* of the weights. The algorithms we present here work for any semiring.

[2] Composition can be defined for any semiring $(K, +, \cdot)$, and the transductions we are considering here can be defined in terms of formal power series [3, 6].

M. Mohri, F. Pereira and M. Riley

## 4 Efficient Composition of Weighted Finite-State Transducers

Composition is the key operation on weighted transducers. The composition algorithm in the weighted case is related to the standard unweighted transducer composition and acceptor intersection algorithms, but in general weighted $\epsilon$-transitions complicate matters, as we will see below. We sketch here an efficient composition algorithm for general weighted transducers.

In general, the input or output label of a transducer transition may be the symbol $\epsilon$ representing a *null* move. A null input label indicates that no symbol needs to be consumed when traversing the transition, while a null output label indicates that no symbol is output when traversing the transition. Null labels are needed because input and output strings do not always have the same length (for instance, a word sequence is much shorter than the corresponding phonetic transcription). Null labels are also a convenient way of delaying inputs or outputs, which may have important computational effects. In the weighted finite-state transducers we use in speech recognition, transitions with null labels may also have a weight.

The presence of null labels makes the composition operation for weighted transducers more delicate than that for unweighted transducers. The problem is illustrated with the composition of the two transducers $A$ and $B$ shown in Figures 3a and 3b. Transducer $A$ has output null transitions, while transducer $B$ has input null transitions. To help understand how these null transitions interact, we show also two derived transducers $A'$ and $B'$ in Figures 3c and 3d. In $A'$, the output null transitions are labeled $\epsilon_2$, and the corresponding null moves in $B'$ are explicitly marked as self-transitions with input label $\epsilon_2$. Likewise, the input null transitions of $B$ are labeled with $\epsilon_1$ in $B'$, and the corresponding self-transitions in $A'$ have output label $\epsilon_1$. Any transition in the composition of $A$ and $B$ has a corresponding transition in the composition of $A'$ and $B'$, but whereas an output null label in $A$ or an input null label in $B$ corresponds to staying in the same state on the other transducer, in the composition of $A'$ and $B'$ the corresponding transition is made from a pair of transitions with matching $A$-output and $B$-input labels $\epsilon_i$.

Figure 4 shows all of the possible transitions in the composition of $A$ and $B$, subscripted with the corresponding matching pair of labels in $A'$ and $B'$. Each path from $(1,1)$ to $(4,3)$ corresponds to a distinct way of using the $\epsilon$-transitions of the two machines depending on the order in which the null moves are taken. However, for composition to be correct in the weighted case, no more than one of those paths should be kept in the result, or else the total weight would be added in as many times as the number of distinct successful paths.

To keep only one of those paths, one can insert a *filter* between $A$ and $B$ (more precisely, between $A'$ and $B'$) that removes the redundant paths. Interestingly, the filter itself can be represented as a finite-state transducer. Filters of different forms are possible, but the one shown in Figure 5 leads in many cases to the fewest transitions in the result, and often to better time efficiency. (The symbol $x$ represents any element of the alphabet of the two transducers.)

The filter can be understood in the following way: as long as the output of $A$ matches the input of $B$, one can move forward on both and stay at state 0. If there is an $\epsilon$-transition in $A$, one can move forward in $A$ (only) and then repeat this operation (state 1) until a possible match occurs which would lead to the state 0 again. Similarly, if there is an $\epsilon$-transition in $B$, one can move forward in $B$ (only) and then repeat this operation (state 2) until a possible match occurs which would lead to the state 0.

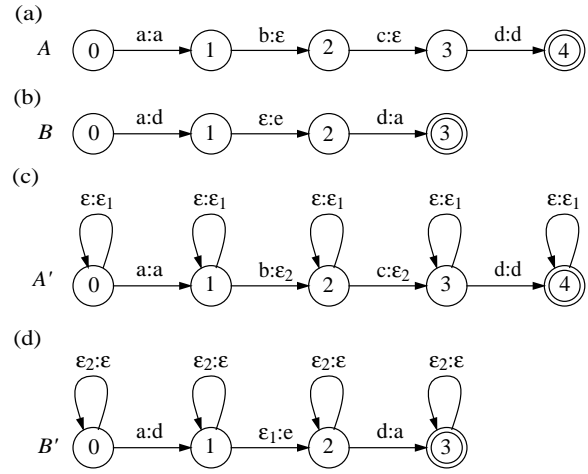

**Figure 3.** Transducers with $\epsilon$ Labels

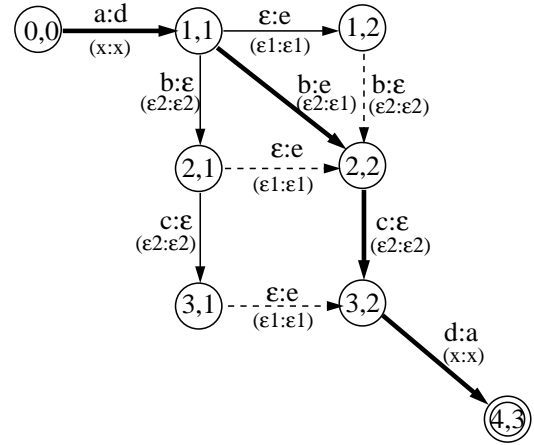

**Figure 4.** Composition with marked $\epsilon$'s

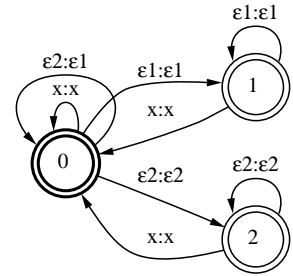

**Figure 5.** Filter Transducer

A crucial algorithmic advantage of transducer composition is that it can be easily computed on the fly. We developed a fully general *lazy* composition algorithm, which creates on demand, and optionally saves for reuse, just those states and arcs of the composed transducer that are required in a particular recognition run, for instance, those required for paths within the given beam width from the best path in a beam-search procedure. We can thus use the lazy composition algorithm as a subroutine in a standard Viterbi decoder to combine on-the-fly a language model, a multi-pronunciation lexicon with corpus-derived pronunciation probabilities, and a context-dependency transducer. The external interface to composed transducers does not distinguish between lazy and precompiled compositions, so the decoder algorithm is the same as for an explicit network.

## 5 Determinization and Minimization of Weighted Automata

Not all weighted automata can be determinized, but weighted automata useful in practice often admit determinization. As in the unweighted case, minimization can be applied to determinized automata. We cannot describe here the theoretical conditions for an automaton to be *subsequentiable*, that is determinizable, or the actual determinization and minimization algorithms for weighted automata. However, these algorithms can play an important role in speech recognition, because they allow us to reduce the size of intermediate compositions and thus save time and space.

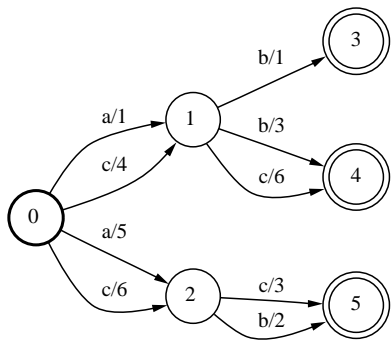

**Figure 6.** Weighted automaton $\alpha_1$.

Figures 6-8 illustrate these algorithms in a simple case. Figure 6 represents a weighted automaton $\alpha_1$. Automaton $\alpha_1$ is not deterministic since, for instance, several transitions with the same input label $a$ leave state 0. Similarly, several transitions with the same input label $b$ leave the state 1.

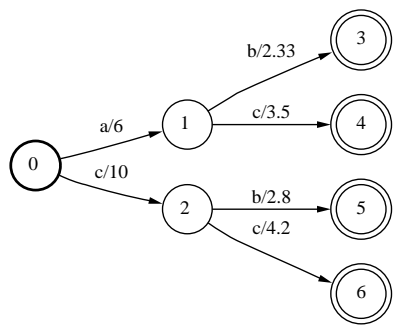

**Figure 7.** Determinized weighted automaton $\alpha_2$.

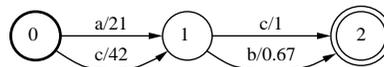

**Figure 8.** Minimized weighted automaton $\alpha_3$.

Automaton $\alpha_1$ can however be determinized.[3] The automaton $\alpha_2$ that results by determinization is shown in figure 7. Clearly $\alpha_2$ is less redundant, since it admits only one path for each input string.

As in the unweighted case, a deterministic weighted automaton can be minimized giving an equivalent deterministic weighted automaton with the fewest number of states.[4] Figure 8 represents the automaton $\alpha_3$ resulting by minimization. It is yes more compact than $\alpha_1$ and it can still be used efficiently because it is deterministic.

## 6 Conclusion

We sketched the application of weighted automata in speech recognition and some of the main algorithms that support it. Weighted finite-state automata can also be used in text-based applications such as the segmentation of Chinese text [13] and text indexation [9].

## Acknowledgments

Hiyan Alshawi, Adam Buchsbaum, Emerald Chung, Don Hindle, Andrej Ljolje, Steven Phillips and Richard Sproat have commented extensively on these ideas, tested many versions of our algorithms, and contributed a variety of improvements. Details of our joint work and their own separate contributions in this area will be presented elsewhere.

---

[3] Notice that if you consider each (input label, weight) pair as a single label applying classical determinization is of no help since the pairs are all distinct.

[4] This minimization algorithm is more general than just applying the classical minimization algorithm to (input label, weight) pairs because weights may need to be moved to make some mergings possible.

M. Mohri, F. Pereira and M. Riley

M. Mohri, F. Pereira and M. Riley